\documentclass[sigconf]{acmart}
\usepackage{booktabs}   
\usepackage{wasysym}   
\usepackage{enumitem}
\usepackage{multirow}
\usepackage{wrapfig}
\AtBeginDocument{%
  }

\copyrightyear{2026}
\acmYear{2026}
\setcopyright{cc}
\setcctype{by}
\acmConference[ETRA '26]{2026 Symposium on Eye Tracking Research and Applications}{June 01--04, 2026}{Marrakesh, Morocco}
\acmBooktitle{2026 Symposium on Eye Tracking Research and Applications (ETRA '26), June 01--04, 2026, Marrakesh, Morocco}
\acmDOI{10.1145/3797246.3806796}
\acmISBN{979-8-4007-2519-7/2026/06}

\begin{document}

\title{GazeVaLM: A Multi-Observer Eye-Tracking Benchmark for Evaluating Clinical Realism in AI-Generated X-Rays}








\author{David Wong$^*$}
\affiliation{
  \institution{Northwestern University}
  \city{Evanston}
  \state{IL}
  \country{United States}
}

\author{Zeynep Isik$^*$}
\affiliation{
  \institution{Northwestern University}
  \city{Evanston}
  \state{IL}
  \country{United States}
}

\author{Bin Wang}
\affiliation{
  \institution{Northwestern University}
  \city{Evanston}
  \state{IL}
  \country{USA}
}

\author{Marouane Tliba}
\affiliation{
  \institution{Université Sorbonne Paris Nord}
  \city{Paris}
  \state{}
  \country{France}
}

\author{Gorkem Durak}
\affiliation{
  \institution{Northwestern University}
  \city{Chicago}
  \state{IL}
  \country{USA}
}

\author{Elif Keles}
\affiliation{
  \institution{Northwestern University}
  \city{Chicago}
  \state{IL}
  \country{USA}
}

\author{Halil Ertugrul Aktas}
\affiliation{
  \institution{Northwestern University}
  \city{Chicago}
  \state{IL}
  \country{USA}
}

\author{Aladine Chetouani}
\affiliation{
  \institution{Université Sorbonne Paris Nord}
  \city{Paris}
  \state{}
  \country{France}
}

\author{Cagdas Topel}
\affiliation{
  \institution{Northwestern University}
  \city{Evanston}
  \state{IL}
  \country{USA}
}

\author{Nicolo Gennaro}
\affiliation{
  \institution{Northwestern University}
  \city{Evanston}
  \state{IL}
  \country{USA}
}

\author{Camila Lopes Vendrami}
\affiliation{
  \institution{Northwestern University}
  \city{Chicago}
  \state{Illinois}
  \country{United States}
}

\author{Tugce Agirlar Trabzonlu}
\affiliation{
  \institution{Northwestern University}
  \city{Evanston}
  \state{IL}
  \country{USA}
}

\author{Amir Ali Rahsepar}
\affiliation{
  \institution{Northwestern University}
  \city{Chicago}
  \state{Illinois}
  \country{United States}
}

\author{Laetitia Perronne}
\affiliation{
  \institution{Northwestern University}
  \city{Evanston}
  \state{IL}
  \country{USA}
}

\author{Matthew Antalek}
\affiliation{
  \institution{Northwestern University}
  \city{Chicago}
  \state{IL}
  \country{USA}
}

\author{Onural Ozturk}
\affiliation{
  \institution{Northwestern University}
  \city{Evanston}
  \state{IL}
  \country{USA}
}

\author{Gokcan Okur}
\affiliation{
  \institution{Loyola University Chicago}
  \city{Chicago}
  \state{IL}
  \country{USA}
}

\author{Andrew C. Gordon}
\affiliation{
  \institution{Northwestern University}
  \city{Chicago}
  \state{IL}
  \country{USA}
}

\author{Ayis Pyrros}
\affiliation{
  \institution{DuPage Medical Group}
  \city{Chicago}
  \state{Illinois}
  \country{United States}
}

\author{Frank H. Miller}
\affiliation{
  \institution{Northwestern University}
  \city{Chicago}
  \state{IL}
  \country{USA}
}

\author{Amir Borhani}
\affiliation{
  \institution{Northwestern University}
  \city{Chicago}
  \state{IL}
  \country{USA}
}

\author{Hatice Savas}
\affiliation{
  \institution{Northwestern University}
  \city{Chicago}
  \state{IL}
  \country{USA}
}

\author{Eric Hart}
\affiliation{
  \institution{Northwestern University}
  \city{Chicago}
  \state{Illinois}
  \country{United States}
}

\author{Elizabeth Krupinski}
\affiliation{
  \institution{Emory University}
  \city{Atlanta}
  \state{GA}
  \country{USA}
}

\author{Ulas Bagci}
\affiliation{
  \institution{Northwestern University}
  \city{Chicago}
  \state{IL}
  \country{USA}
}

\renewcommand{\shortauthors}{}

\begin{abstract}
We introduce \textbf{GazeVaLM}, a public eye-tracking dataset for studying clinical perception during chest radiograph authenticity assessment. The dataset comprises 960 gaze recordings from 16 expert radiologists interpreting 30 real and 30 synthetic chest X-rays (generated by diffusion based generative AI) under two conditions: diagnostic assessment and real-fake classification (Visual Turing test). For each image–observer pair, we provide raw gaze samples, fixation maps, scanpaths, saliency density maps, structured diagnostic labels, and authenticity judgments. We extend the protocol to 6 state-of-the-art multimodal LLMs, releasing their predicted diagnoses, authenticity labels, and confidence scores under matched conditions — enabling direct human–AI comparison at both decision and uncertainty levels. We further provide analyses of gaze agreement, inter-observer consistency, and benchmarking of radiologists versus LLMs in diagnostic accuracy and authenticity detection. GazeVaLM supports research in gaze modeling, clinical decision-making, human–AI comparison, generative image realism assessment, and uncertainty quantification. By jointly releasing visual attention data, clinical labels, and model predictions, we aim to facilitate reproducible research on how experts and AI systems perceive, interpret, and evaluate medical images. The dataset is available at \url{https://huggingface.co/datasets/davidcwong/GazeVaLM}.
\end{abstract}





\maketitle

\section{Introduction}
Synthetic medical image generation has emerged as a critical tool for overcoming the data scarcity, privacy constraints, and class imbalance that limit clinical AI development~\cite{litjens2017survey, esteva2021deep}. Generative adversarial networks~\cite{NIPS2014_5ca3e9b1}, variational autoencoders~\cite{rezende2014stochasticbackpropagationapproximateinference, kingma2022autoencodingvariationalbayes}, and diffusion models~\cite{NEURIPS2020_4c5bcfec} can now produce chest radiographs of striking visual fidelity~\cite{chen2021synthetic, pengfei2024addressing}. However, the quality of synthetic images is typically evaluated using computational metrics such as FID and IS, which measure distributional similarity but do not capture whether images are \emph{clinically authentic}~\cite{chuquicusma2018fool, theis2015note, borji2019pros}. An image may score well on standard metrics while containing subtle artifacts that an experienced radiologist would immediately recognize as unrealistic~\cite{jungImageBasedGenerativeArtificial2024}. This gap between computational evaluation and clinical perception represents a fundamental limitation in how synthetic medical images are currently validated.

Bridging this gap requires understanding how clinicians \emph{visually} engage with synthetic images. Eye tracking provides a direct, non-invasive window into the perceptual and cognitive processes underlying radiological interpretation~\cite{drew2013,krupinski2010}. Expert gaze behavior encodes where clinicians attend, in what order, and for how long---revealing the spatial reasoning process that computational metrics cannot capture. The Visual Turing Test (VTT)~\cite{chuquicusma2018}, which asks experts to classify images as real or synthetic, operationalizes clinical authenticity in a measurable experimental paradigm. Yet no existing dataset captures \emph{how experts visually process} both real and synthetic medical images during such assessments.

We introduce \textbf{GazeVaLM}, the first multi-observer eye-tracking benchmark for evaluating clinical realism in AI-generated chest radiographs. GazeVaLM addresses three gaps simultaneously: (1)~it captures expert gaze during both diagnostic interpretation and authenticity assessment, enabling comparison of visual strategies across tasks; (2)~it pairs every synthetic image with its real counterpart (generated from identical clinical reports), providing controlled comparisons; and (3)~it extends the same experimental protocol to six state-of-the-art multimodal LLMs, enabling direct human--AI comparison on both diagnostic accuracy and authenticity detection.

\begin{figure*}[h]
\center
    \includegraphics[width=0.7\linewidth]{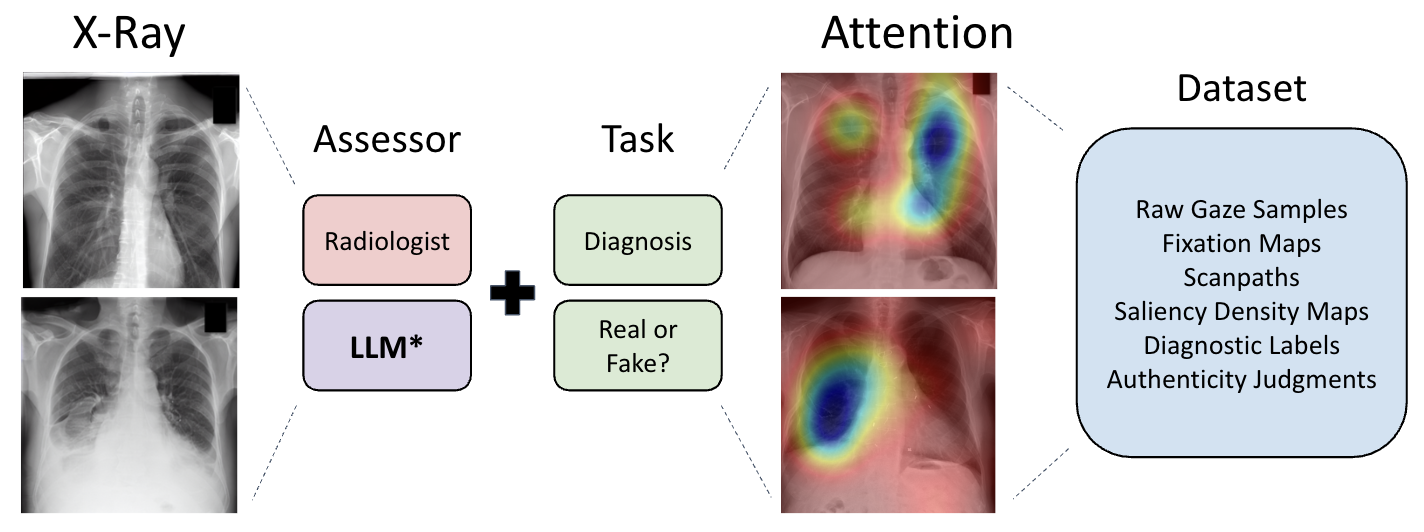}
    \caption{Overview of the proposed dataset pipeline, which introduces has two different assessors undertake two different tasks to produce an assortment of gaze data.}
    \label{framework}
    \vspace{-13pt}
\end{figure*}

Our key contributions are:
\begin{itemize}
    \item A publicly available dataset of 960 gaze recordings from 16 expert radiologists interpreting 30 real and 30 synthetic chest X-rays under two experimental conditions, with raw gaze samples, fixation maps, scanpaths, saliency density maps, diagnostic labels, and authenticity judgments.
    \item A matched evaluation of six multimodal LLMs on identical stimuli and tasks, with predicted diagnoses, authenticity labels, and confidence scores.
    \item Empirical analyses revealing that pupillometric measures provide a robust implicit marker of perceived image authenticity, with real images consistently eliciting larger pupil size and weaker constriction than synthetic counterparts.
\end{itemize}


\paragraph{Related Work.} Eye tracking has a rich history in radiology research, dating to Kundel and Nodine's foundational work on visual search during chest X-ray interpretation~\cite{kundel1975}. Recent work has used gaze data to supervise deep learning models. GazeSAM~\cite{wang2024gazesam} uses fixation points as prompts for interactive segmentation. EGMA~\cite{ma2024eye} aligns image--text representations using gaze. GazeGNN~\cite{wang2024gazegnn} uses gaze-guided graph neural networks for chest X-ray classification. These systems demonstrate that gaze data is a valuable training signal; GazeVaLM extends this by providing gaze data captured during a novel task---authenticity assessment---that could inform future gaze-supervised models for synthetic image quality evaluation.

While the gaze-guided AI in medical imaging is getting more attraction as time goes on, available benchmarking and datasets are extremely limited. Current resources, such as REFLACX \cite{BigolinLanfredi2022REFLACX}, EGD-CXR \cite{karargyris2020eye}, MIMIC-EYE \cite{mimic-eye}, CT-ScanGaze \cite{pham2025ct}, are useful datasets, and the field is active and growing with an unmet need for actions dataset and benchmarking. However, none of these datasets provide a visual Turing test and assess clinical realism.

\section{Dataset - Synthetic Chest X-ray Generation} \label{gen}
\begin{figure*}[ht]                                                                                      
      \centering                                            
      \includegraphics[width=0.8\textwidth]{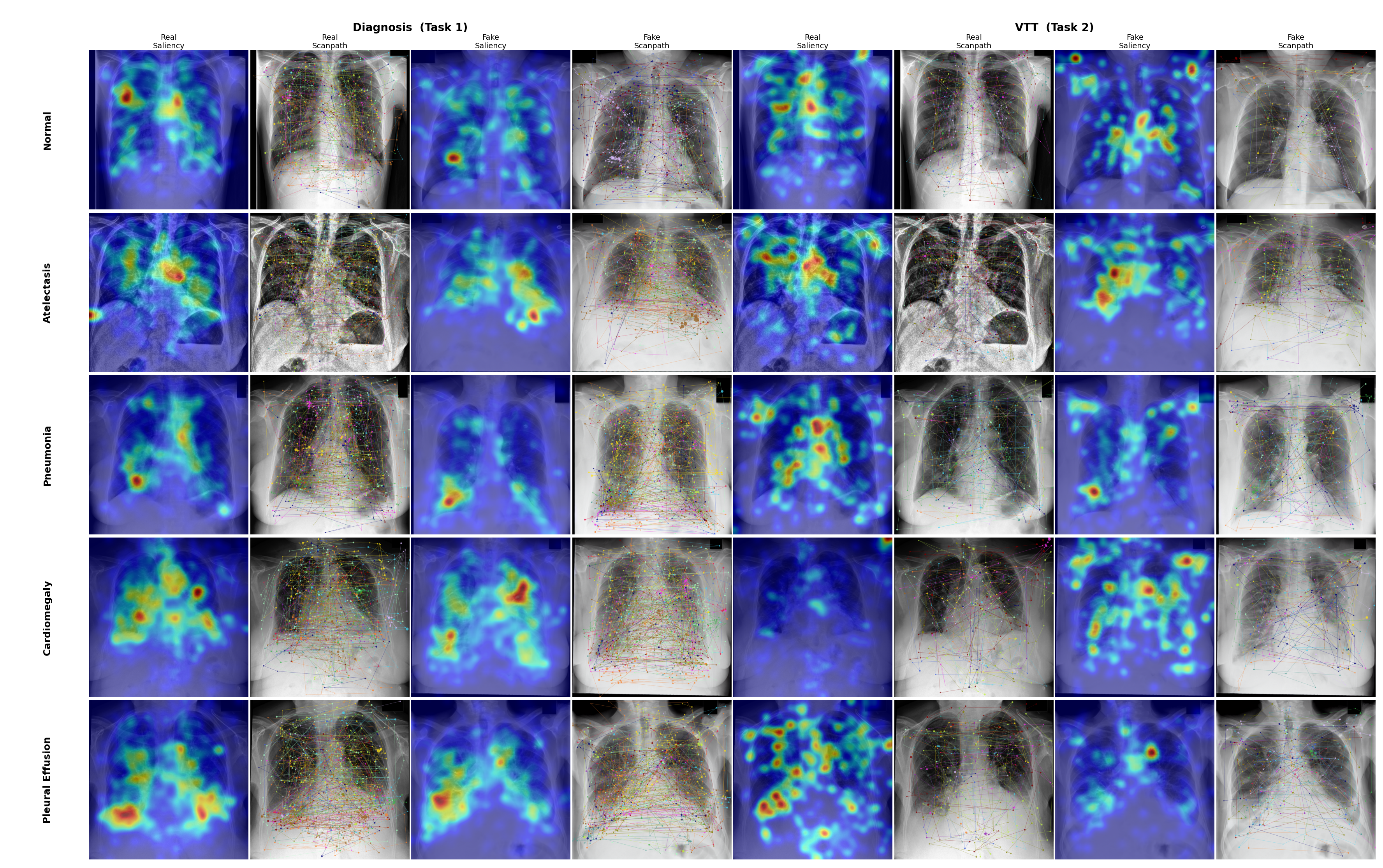}                                       
      \caption{Gaze overlays organized by radiograph diagnosis for Diagnosis Task (Task~1)
      and Visual Turing Test (Task~2), across real X-rays and AI-generated images.                        
      Each row corresponds to a pathology category.
      Columns alternate between saliency maps (duration-weighted Gaussian heatmaps,
      $\sigma=25$\,px) and scanpath overlays (up to 16 observers)}
      \label{fig:gaze_grid}
\end{figure*}

The GazeVaLM dataset is constructed through a four-stage pipeline (Fig.\ref{framework}): synthetic image generation, multi-observer eye-tracking study, gaze data processing, and LLM evaluation.
Synthetic chest X-rays were generated using RoentGen~\cite{bluethgenVisionlanguageFoundationModel2024}, a vision-language diffusion model conditioned on free-text radiology reports. We selected reports from MIMIC-CXR~\cite{johnson2019mimic} meeting RoentGen's 77-token input limit, ensuring each synthetic image has a paired real counterpart with identical report content. Following generation, an independent board-certified radiologist reviewed all outputs and excluded images exhibiting significant unrealistic features (e.g., gross anatomical distortions, text artifacts). Thirty images were randomly selected from the remaining pool, yielding a final stimulus set of 30 real and 30 synthetic chest X-rays spanning five pathology categories: normal, atelectasis, cardiomegaly, pleural effusion, and pneumonia. 
Some example of synthetic chest X-rays overlaid with scan-path are showed in Fig.\ref{fig:gaze_grid}.


\subsection{Eye-Tracking Study}\label{eyetrackingstudy}

\subsubsection{Participants}
Sixteen board-certified radiologists (5 women, 11 men) participated in the study. Experience ranged from 0--4 years ($n=2$), 5--9 years ($n=6$), 10--19 years ($n=5$), to 20+ years ($n=3$), spanning body imaging, cardiothoracic radiology, interventional radiology, neuroradiology, and musculoskeletal radiology (Table~\ref{YoE}). All participants had normal or corrected-to-normal vision.

\begin{table}[h]
\centering
\footnotesize

\begin{minipage}{0.48\textwidth}
\centering
\caption{VTT task accuracy and confidence scores of LLMs and comparison with human accuracy}
\begin{tabular}{lcc}
\toprule
 & \textbf{Accuracy (\%)} & \textbf{Confidence} \\
\midrule
\textbf{LLM} &  &  \\
\midrule
GPT 5.2 Thinking & 71.667 & 3.000 \\
Gemini 3 Pro & 66.667 & 3.833 \\
Claude Sonnet 4.5 Thinking & 65.000 & 2.833 \\
Grok 4.1 & 58.333 & 3.700 \\
Llama 4 & 55.000 & 3.367 \\
DeepSeek 3.2 & 50.000 & 3.100 \\
\midrule
\textbf{Human Expert} & 80.104 & - \\
\bottomrule
\end{tabular}
\label{tab:llm_vtt}
\end{minipage}
\hfill

\begin{minipage}{0.48\textwidth}
\centering
\caption{Radiologists listed by specialty and years of experience. Radiologists with multiple subspecialities are listed multiple times.}
\resizebox{\textwidth}{!}{
\begin{tabular}{|c|c|c|c|c|}
\hline
\multirow{2}{*}{Specialty} & \multicolumn{4}{c|}{Years of Experience} \\
\cline{2-5}
 & 0--9 & 10--19 & 20+ & Total \\
\hline
Body Imaging & 3 & 3 & 2 & 8 \\
\hline
Cardiothoracic Imaging & 4 & 2 & 1 & 7 \\
\hline
Musculoskeletal Imaging & 1 & 0 & 1 & 2 \\
\hline
Neuroradiology & 0 & 0 & 1 & 1 \\
\hline
Interventional Radiology & 1 & 1 & 0 & 2 \\
\hline
\end{tabular}}
\label{YoE}
\end{minipage}

\end{table}

\subsubsection{Stimuli and Eye Tracker Setup}
The final stimulus set comprised 30 real chest X-rays drawn from MIMIC-CXR and 30 paired synthetic chest X-rays generated by RoentGen, as described in Sec.~\ref{gen}. Images were combined and presented in a randomized order to each participant.
Eye movements were recorded using EyeLink 1000 Plus eye tracker. Gaze positions were computed from a 13-point calibration procedure validated by the EyeLink system, which also compensates for blinks, head movements, and other recording artifacts. Fixations were defined as periods during which saccade velocity remained below 30°/s and acceleration below $8000°/s^2$ \cite{cornelissen2002eyelink}.

\subsubsection{Experimental Tasks}
Each radiologist completed two tasks on the same 60 images (30 real, 30 synthetic), presented in randomized order.

\textit{Task~1: Diagnostic Assessment.} Radiologists verbally described pathologies and findings for each image. Crucially, they were not informed that synthetic images were present. This task captures naturalistic diagnostic gaze behavior uncontaminated by authenticity awareness.

\textit{Task~2: Visual Turing Test (VTT).} Radiologists classified each image as real or synthetic, reporting the visual cues underlying their decision. They were informed of the presence of synthetic images but not their proportion. Task~2 was conducted after an average washout interval of 10 days to reduce carryover effects from prior exposure.

This dual-task design is a key feature of GazeVaLM: by comparing gaze patterns for the same images under different task demands (diagnostic vs.\ authenticity), the dataset enables analysis of how awareness of synthetic images alters visual search strategies.


Raw gaze samples were processed into fixation sequences using a velocity-based dispersion algorithm. Fixation maps were computed as duration-weighted Gaussian heatmaps. Scanpaths were extracted as ordered sequences of fixation coordinates and durations for each observer and trial. Examples are shown in Fig.~\ref{fig:gaze_grid}.

\subsection{LLM Evaluation}\label{llms}
To complement the human gaze study and enable direct comparison between clinical experts and AI systems, we extend the same experimental protocol to state-of-the-art multimodal LLMs. As these models are increasingly deployed in medical image interpretation tasks, understanding how their diagnostic judgments and authenticity assessments align with or different from the expert radiologists opinions. By subjecting LLMs to identical stimuli and tasks, we obtain a controlled basis for benchmarking human and AI perception of both real and synthetic chest X-rays.

Six multimodal LLMs were evaluated using the ChatHub interface, which supports simultaneous interaction with multiple models using a single prompt: GPT 5.2 Thinking~\cite{singh2025openai}, Llama 4~\cite{adcock2026llama}, DeepSeek 3.2 ~\cite{liu2024deepseek}, Gemini 3 Pro, Claude Sonnet 4.5 Thinking~\cite{claude}, and Grok 4.1~\cite{grok}. A new chat session was initialized for every image to prevent learning effects across trials.
\begin{table*}[h]
\centering
\scriptsize
\setlength{\tabcolsep}{3pt}
\renewcommand{\arraystretch}{0.95}

\caption{Gaze statistics per task and image type (mean$\pm$SD across participants).
($N_{\text{Task1}}=21$, $N_{\text{Task2}}=18$).}
\label{tab:gaze_stats}

\begin{tabular}{lcccccc}
\toprule
 & \multicolumn{3}{c}{\textbf{Diagnosis (Task 1)}} & \multicolumn{3}{c}{\textbf{VTT (Task 2)}} \\
\cmidrule(lr){2-4}\cmidrule(lr){5-7}
Metric & All & Real & Fake & All & Real & Fake \\
\midrule

\multicolumn{7}{l}{\textit{Fixation}} \\
Fixations per trial & 75.5$\pm$45.8 & 73.3$\pm$43.0 & 77.6$\pm$49.0 & 26.8$\pm$17.0 & 26.4$\pm$18.4 & 27.2$\pm$15.9 \\
Fixation duration (ms) & 274.2$\pm$34.9 & 276.2$\pm$36.7 & 272.2$\pm$34.2 & 272.8$\pm$33.6 & 273.9$\pm$36.5 & 271.8$\pm$32.2 \\

\multicolumn{7}{l}{\textit{Saccade / Scanpath}} \\
Saccade amplitude (deg) & 6.41$\pm$1.55 & 6.49$\pm$1.55 & 6.32$\pm$1.55 & 6.89$\pm$1.26 & 7.01$\pm$1.28 & 6.76$\pm$1.27 \\
Scanpath length (deg) & 381.9$\pm$191.7 & 374.8$\pm$182.8 & 389.0$\pm$202.6 & 150.6$\pm$83.5 & 150.4$\pm$92.7 & 150.7$\pm$77.6 \\

\multicolumn{7}{l}{\textit{Velocity}} \\
Mean saccade vel.\ (deg/s) & 159.3$\pm$24.6 & 161.2$\pm$24.7 & 157.4$\pm$24.8 & 169.9$\pm$26.7 & 172.1$\pm$26.3 & 167.6$\pm$27.5 \\
Peak saccade vel.\ (deg/s) & 264.9$\pm$48.7 & 268.6$\pm$49.7 & 261.2$\pm$48.3 & 286.5$\pm$63.8 & 291.1$\pm$63.3 & 281.7$\pm$64.6 \\

\multicolumn{7}{l}{\textit{Pupil}} \\
Mean pupil (a.u.) & 252.3$\pm$94.3 & 275.4$\pm$104.4 & 229.4$\pm$85.0 & 221.7$\pm$76.4 & 237.4$\pm$78.7 & 205.9$\pm$75.5 \\
Pupil dilation (a.u.) & $-7.64\pm13.46$ & $+2.95\pm14.96$ & $-18.19\pm16.90$ & $-11.73\pm14.87$ & $-5.00\pm15.96$ & $-18.51\pm16.79$ \\
Pupil variability (a.u.) & 36.91$\pm$17.73 & 40.47$\pm$19.33 & 33.37$\pm$16.50 & 28.88$\pm$12.14 & 30.90$\pm$13.53 & 26.86$\pm$11.06 \\

\multicolumn{7}{l}{\textit{Timing}} \\
Viewing time (s) & 26.50$\pm$16.94 & 25.59$\pm$16.01 & 27.40$\pm$18.03 & 8.85$\pm$6.56 & 8.72$\pm$7.10 & 9.00$\pm$6.17 \\

\bottomrule
\end{tabular}

\end{table*}

Following the same structure as the radiologist study, the real and synthetic images were combined and presented in randomized order. In Task 1 (Diagnostic Assessment), models were conditioned to act as board-certified radiologists and prompted to provide findings, impressions, and a confidence rating (1 = very low, 2 = low, 3 = high, 4 = very high) for each image, returned as a structured JSON. Note that, in this stage, models were not informed that synthetic images were present. In Task 2 (VTT), models were informed that some images were synthetic and asked to classify each image as real or fake, report the visual cues and features underlying their decision, and return a confidence rating, again in JSON format.

\section{Analysis}
\subsection{Gaze statistics across tasks and image types}
Table~\ref{tab:gaze_stats} reports gaze and pupil metrics for Diagnosis (Task~1) and VTT (Task~2), averaged across participants and split by image authenticity (Real vs.\ Fake). Fixation duration is measured in milliseconds (ms); saccade amplitude and scanpath length in degrees of visual angle (deg); saccade velocities in deg/s; pupil size and dilation in EyeLink arbitrary units (a.u.); and viewing time in seconds (s).

Fixation-level metrics are largely driven by viewing time. Diagnosis trials contain about 2.8$\times$ more fixations than VTT (75.5 vs.\ 26.8), in line with the 3.0$\times$ difference in viewing time (26.5 vs.\ 8.9\,s), whereas fixation duration is almost identical across tasks (274.2 vs.\ 272.8\,ms). This indicates that observers maintain similar dwell times per fixation and simply sample the image for longer in Diagnosis. Within Diagnosis, fake images attract slightly more fixations and marginally shorter dwell times than real (77.6 vs.\ 73.3 fixations; 272.2 vs.\ 276.2\,ms), a small effect that largely disappears in VTT.

Saccade and scanpath metrics reflect changes in sampling scale. VTT elicits somewhat larger and faster saccades than Diagnosis (6.89 vs.\ 6.41\,deg; 169.9 vs.\ 159.3\,deg/s; 286.5 vs.\ 264.9\,deg/s), consistent with a slightly more global exploration strategy under time pressure. Scanpath length scales with viewing time (381.9 vs.\ 150.6\,deg), suggesting a roughly constant exploration rate across tasks. Real images show slightly larger saccade amplitudes than fake in both tasks, but these differences are modest compared to the task effect.

Pupil metrics provide the clearest signature of image authenticity. Mean pupil size is higher for real than fake images in both tasks (Diagnosis: 275.4 vs.\ 229.4\,a.u.; VTT: 237.4 vs.\ 205.9\,a.u.), and baseline-corrected dilation shows a consistent pattern: real images remain close to baseline (+2.95\,a.u.\ in Diagnosis; $-5.00$\,a.u.\ in VTT), whereas fake images induce stronger constriction ($-18.19$ and $-18.51$\,a.u.). Pupil variability is also higher for real than fake images in both tasks. Overall, these statistics indicate that task design primarily controls exploration duration and scale, while pupil-based measures are the most sensitive markers of perceived authenticity.

\subsection{LLM Performance on VTT}
Table~\ref{tab:llm_vtt} reports VTT accuracy and confidence scores across six LLMs alongside human expert performance. ChatGPT-5.2 Thinking achieves the highest accuracy among models (71.7\%), while DeepSeek 3.2 performs at chance level (50.0\%). All six LLMs fall below human expert accuracy (80.1\%), despite reporting consistently high confidence scores, suggesting a systematic overconfidence in authenticity judgments relative to actual performance.

The complete GazeVaLM dataset structure—including radiologist gaze recordings for both tasks, fixation and saliency maps, stimuli images, and LLM evaluation outputs—is available on our Hugging Face repository, where the full directory organization and file descriptions are documented.


\section{Discussion and Conclusion}
\label{sec:conclusion}
We presented \textbf{GazeVaLM}, the first multi-observer eye-tracking benchmark designed to evaluate the clinical realism of AI-generated chest radiographs. By capturing gaze data from 16 expert radiologists across two complementary tasks---diagnostic interpretation and Visual Turing Test---and extending the same protocol to six multimodal LLMs, GazeVaLM enables controlled comparison of human and AI perception of synthetic medical images. Our analyses reveal that pupillometric measures provide a robust implicit signature of perceived authenticity, that task design primarily controls exploration strategy while pupil dynamics track image realism, and that current LLMs exhibit systematic overconfidence in authenticity judgments. The joint release of gaze recordings, clinical labels, and LLM predictions establishes a reproducible foundation for research in gaze modeling, human--AI alignment, generative image evaluation, and uncertainty quantification.
Several limitations should be acknowledged. The dataset is restricted to frontal chest radiography generated by a single diffusion model (RoentGen), limiting generalizability to other modalities, views, and generative architectures. The cohort of 16 radiologists, while diverse in experience and subspecialty, is drawn from a single institution. The stimulus set of 60 images (30 real, 30 synthetic) is modest in scale, though the multi-observer design yields 960 recordings. Future work should extend the benchmark to additional imaging modalities, multiple generative architectures, and multi-site observer populations.

\section{Privacy \& Ethics Statement}
This dataset enables research on how expert radiologists and LLM perceive real and synthetic medical images, which will improve the reliability of AI models used in healthcare. Synthetic medical images may be misused to generate misleading clinical data or produce models that fail under real-world conditions if it is not carefully evaluated. By releasing gaze data and evaluation results, this work aims to promote the responsible development of AI in medical imaging.

\bibliographystyle{ACM-Reference-Format}
\bibliography{sample-base}

@article{bluethgenVisionlanguageFoundationModel2024,
  title = {A Vision-Language Foundation Model for the Generation of Realistic Chest {{X-ray}} Images},
  author = {Bluethgen, Christian and Chambon, Pierre and Delbrouck, Jean-Benoit and {van der Sluijs}, Rogier and Po{\l}acin, Ma{\l}gorzata and Zambrano Chaves, Juan Manuel and Abraham, Tanishq Mathew and Purohit, Shivanshu and Langlotz, Curtis P. and Chaudhari, Akshay S.},
  year = {2024},
  month = aug,
  journal = {Nature Biomedical Engineering},
  issn = {2157-846X},
  doi = {10.1038/s41551-024-01246-y},
  langid = {english},
  pmcid = {PMC11861387},
  pmid = {39187663}
}

@article{johnson2019mimic,
  title={MIMIC-CXR, a de-identified publicly available database of chest radiographs with free-text reports},
  author={Johnson, Alistair EW and Pollard, Tom J and Berkowitz, Seth J and Greenbaum, Nathaniel R and Lungren, Matthew P and Deng, Chih-ying and Mark, Roger G and Horng, Steven},
  journal={Scientific data},
  volume={6},
  number={1},
  pages={317},
  year={2019},
  publisher={Nature Publishing Group UK London}
}

@article{cornelissen2002eyelink,
  title={The Eyelink Toolbox: eye tracking with MATLAB and the Psychophysics Toolbox},
  author={Cornelissen, Frans W and Peters, Enno M and Palmer, John},
  journal={Behavior Research Methods, Instruments, \& Computers},
  volume={34},
  number={4},
  pages={613--617},
  year={2002},
  publisher={Springer}
}

@inproceedings{NIPS2014_5ca3e9b1,
 author = {Goodfellow, Ian and Pouget-Abadie, Jean and Mirza, Mehdi and Xu, Bing and Warde-Farley, David and Ozair, Sherjil and Courville, Aaron and Bengio, Yoshua},
 booktitle = {Advances in Neural Information Processing Systems},
 editor = {Z. Ghahramani and M. Welling and C. Cortes and N. Lawrence and K.Q. Weinberger},
 pages = {},
 publisher = {Curran Associates, Inc.},
 title = {Generative Adversarial Nets},
 url = {https://proceedings.neurips.cc/paper_files/paper/2014/file/5ca3e9b122f61f8f06494c97b1afccf3-Paper.pdf},
 volume = {27},
 year = {2014}
}

@inproceedings{NEURIPS2020_4c5bcfec,
 author = {Ho, Jonathan and Jain, Ajay and Abbeel, Pieter},
 booktitle = {Advances in Neural Information Processing Systems},
 editor = {H. Larochelle and M. Ranzato and R. Hadsell and M.F. Balcan and H. Lin},
 pages = {6840--6851},
 publisher = {Curran Associates, Inc.},
 title = {Denoising Diffusion Probabilistic Models},
 url = {https://proceedings.neurips.cc/paper_files/paper/2020/file/4c5bcfec8584af0d967f1ab10179ca4b-Paper.pdf},
 volume = {33},
 year = {2020}
}

@misc{kingma2022autoencodingvariationalbayes,
      title={Auto-Encoding Variational Bayes}, 
      author={Diederik P Kingma and Max Welling},
      year={2022},
      eprint={1312.6114},
      archivePrefix={arXiv},
      primaryClass={stat.ML},
      url={https://arxiv.org/abs/1312.6114}, 
}

@misc{rezende2014stochasticbackpropagationapproximateinference,
      title={Stochastic Backpropagation and Approximate Inference in Deep Generative Models}, 
      author={Danilo Jimenez Rezende and Shakir Mohamed and Daan Wierstra},
      year={2014},
      eprint={1401.4082},
      archivePrefix={arXiv},
      primaryClass={stat.ML},
      url={https://arxiv.org/abs/1401.4082}, 
}

@inproceedings{wang2024gazegnn,
  title={Gazegnn: A gaze-guided graph neural network for chest x-ray classification},
  author={Wang, Bin and Pan, Hongyi and Aboah, Armstrong and Zhang, Zheyuan and Keles, Elif and Torigian, Drew and Turkbey, Baris and Krupinski, Elizabeth and Udupa, Jayaram and Bagci, Ulas},
  booktitle={Proceedings of the IEEE/CVF Winter Conference on Applications of Computer Vision},
  pages={2194--2203},
  year={2024}
}

@article{jungImageBasedGenerativeArtificial2024,
  title = {Image-{{Based Generative Artificial Intelligence}} in {{Radiology}}: {{Comprehensive Updates}}},
  author = {Jung, Ha Kyung and Kim, Kiduk and Park, Ji Eun and Kim, Namkug},
  year = {2024},
  month = nov,
  journal = {Korean J Radiol},
  volume = {25},
  number = {11},
  pages = {959--981},
  publisher = {The Korean Society of Radiology},
  issn = {1229-6929}
}

@article{karargyris2020eye,
  title={Eye gaze data for chest x-rays},
  author={Karargyris, Alexandros and Kashyap, Satyananda and Lourentzou, Ismini and Wu, Joy and Tong, Matthew and Sharma, Arjun and Abedin, Shafiq and Beymer, David and Mukherjee, Vandana and Krupinski, Elizabeth and others},
  journal={PhysioNet https://doi. org/10.13026/QFDZ-ZR67},
  year={2020}
}

@article{BigolinLanfredi2022REFLACX,
  title        = {{REFLACX, a dataset of reports and eye-tracking data for localization of abnormalities in chest x-rays}},
  author       = {Bigolin Lanfredi, Ricardo and Zhang, Mingyuan and Auffermann, William F. and Chan, Jessica and Duong, Phuong-Anh T. and Srikumar, Vivek and Drew, Trafton and Schroeder, Joyce D. and Tasdizen, Tolga},
  journal      = {Scientific Data},
  volume       = {9},
  number       = {1},
  pages        = {350},
  year         = {2022},
  doi          = {10.1038/s41597-022-01441-z},
  url          = {https://www.nature.com/articles/s41597-022-01441-z}
}

@article{singh2025openai,
  title={Openai gpt-5 system card},
  author={Singh, Aaditya and Fry, Adam and Perelman, Adam and Tart, Adam and Ganesh, Adi and El-Kishky, Ahmed and McLaughlin, Aidan and Low, Aiden and Ostrow, AJ and Ananthram, Akhila and others},
  journal={arXiv preprint arXiv:2601.03267},
  year={2025}
}

@article{adcock2026llama,
  title={The Llama 4 Herd: Architecture, Training, Evaluation, and Deployment Notes},
  author={Adcock, Aaron and Srivastava, Aayushi and Dubey, Abhimanyu and Jauhri, Abhinav and Pande, Abhinav and Pandey, Abhinav and Sharma, Abhinav and Kadian, Abhishek and Kumawat, Abhishek and Kelsey, Adam and others},
  journal={arXiv preprint arXiv:2601.11659},
  year={2026}
}

@article{liu2024deepseek,
  title={Deepseek-v3 technical report},
  author={Liu, Aixin and Feng, Bei and Xue, Bing and Wang, Bingxuan and Wu, Bochao and Lu, Chengda and Zhao, Chenggang and Deng, Chengqi and Zhang, Chenyu and Ruan, Chong and others},
  journal={arXiv preprint arXiv:2412.19437},
  year={2024}
}

@misc{claude,
  title={System Card:Claude Sonnet 4.5},
  author={{Anthropic}},
  year={2025},
  url={https://www-cdn.anthropic.com/963373e433e489a87a10c823c52a0a013e9172dd.pdf}
}

@misc{grok,
  title={Grok 4.1 Model Card},
  author={{xAI}},
  year={2025},
  url={https://data.x.ai/2025-11-17-grok-4-1-model-card.pdf}
}

@article{litjens2017survey,
  title={A survey on deep learning in medical image analysis},
  author={Litjens, Geert and Kooi, Thijs and Bejnordi, Babak Ehteshami and Setio, Arnaud Arindra Adiyoso and Ciompi, Francesco and Ghafoorian, Mohsen and Van Der Laak, Jeroen Awm and Van Ginneken, Bram and S{\'a}nchez, Clara I},
  journal={Medical image analysis},
  volume={42},
  pages={60--88},
  year={2017},
  publisher={Elsevier}
}

@article{esteva2021deep,
  title={Deep learning-enabled medical computer vision},
  author={Esteva, Andre and Chou, Katherine and Yeung, Serena and Naik, Nikhil and Madani, Ali and Mottaghi, Ali and Liu, Yun and Topol, Eric and Dean, Jeff and Socher, Richard},
  journal={NPJ digital medicine},
  volume={4},
  number={1},
  pages={5},
  year={2021},
  publisher={Nature Publishing Group UK London}
}

@article{pengfei2024addressing,
  title={Addressing medical imaging limitations with synthetic data generation},
  author={Pengfei, G and Dong, Y and Can, Z and Daguang, X},
  journal={NVidia Tech. Blog},
  year={2024}
}

@article{chen2021synthetic,
  title={Synthetic data in machine learning for medicine and healthcare},
  author={Chen, Richard J and Lu, Ming Y and Chen, Tiffany Y and Williamson, Drew FK and Mahmood, Faisal},
  journal={Nature Biomedical Engineering},
  volume={5},
  number={6},
  pages={493--497},
  year={2021},
  publisher={Nature Publishing Group UK London}
}

@inproceedings{chuquicusma2018fool,
  title={How to fool radiologists with generative adversarial networks? A visual turing test for lung cancer diagnosis},
  author={Chuquicusma, Maria JM and Hussein, Sarfaraz and Burt, Jeremy and Bagci, Ulas},
  booktitle={2018 IEEE 15th international symposium on biomedical imaging (ISBI 2018)},
  pages={240--244},
  year={2018},
  organization={IEEE}
}

@article{theis2015note,
  title={A note on the evaluation of generative models},
  author={Theis, Lucas and Oord, A{\"a}ron van den and Bethge, Matthias},
  journal={arXiv preprint arXiv:1511.01844},
  year={2015}
}

@article{borji2019pros,
  title={Pros and cons of GAN evaluation measures},
  author={Borji, Ali},
  journal={Computer vision and image understanding},
  volume={179},
  pages={41--65},
  year={2019},
  publisher={Elsevier}
}

@article{mimic-eye,
  author = {Hsieh, Chihcheng and Ouyang, Chun and Nascimento, Jacinto C and Pereira, Joao and Jorge, Joaquim and Moreira, Catarina},
  title = {{MIMIC-Eye: Integrating MIMIC Datasets with REFLACX and Eye Gaze for Multimodal Deep Learning Applications}},
  journal = {{PhysioNet}},
  year = {2023},
  month = mar,
  note = {Version 1.0.0},
  doi = {10.13026/pc72-as03},
  url = {https://doi.org/10.13026/pc72-as03}
}

@inproceedings{pham2025ct,
  title={CT-ScanGaze: A Dataset and Baselines for 3D Volumetric Scanpath Modeling},
  author={Pham, Trong Thang and Awasthi, Akash and Khan, Saba and Marti, Esteban Duran and Nguyen, Tien-Phat and Vo, Khoa and Tran, Minh and Nguyen, Son and Tran, Cuong and Ikebe, Yuki and others},
  booktitle={Proceedings of the IEEE/CVF International Conference on Computer Vision},
  pages={21732--21743},
  year={2025}
}

@article{drew2013,
  title={The invisible gorilla strikes again: sustained inattentional blindness in expert observers},
  author={Drew, Trafton and V{\~o}, Melissa L-H and Wolfe, Jeremy M},
  journal={Psychological science},
  volume={24},
  number={9},
  pages={1848--1853},
  year={2013},
  publisher={Sage Publications Sage CA: Los Angeles, CA}
}

@article{krupinski2010,
  title={Current perspectives in medical image perception},
  author={Krupinski, Elizabeth A},
  journal={Attention, Perception, \& Psychophysics},
  volume={72},
  number={5},
  pages={1205--1217},
  year={2010},
  publisher={Springer}
}

@INPROCEEDINGS{chuquicusma2018,
  author={Chuquicusma, Maria J. M. and Hussein, Sarfaraz and Burt, Jeremy and Bagci, Ulas},
  booktitle={2018 IEEE 15th International Symposium on Biomedical Imaging (ISBI 2018)}, 
  title={How to fool radiologists with generative adversarial networks? A visual turing test for lung cancer diagnosis}, 
  year={2018},
  volume={},
  number={},
  pages={240-244},
  doi={10.1109/ISBI.2018.8363564}}

@article{kundel1975,
  title={Interpreting chest radiographs without visual search},
  author={Kundel, Harold L and Nodine, Calvin F},
  journal={Radiology},
  volume={116},
  number={3},
  pages={527--532},
  year={1975},
  publisher={The Radiological Society of North America}
}

@inproceedings{wang2024gazesam,
  title={Gazesam: Interactive image segmentation with eye gaze and segment anything model},
  author={Wang, Bin and Aboah, Armstrong and Zhang, Zheyuan and Pan, Hongyi and Bagci, Ulas},
  booktitle={Gaze Meets Machine Learning Workshop},
  pages={254--265},
  year={2024},
  organization={PMLR}
}

@article{ma2024eye,
  title={Eye-gaze guided multi-modal alignment for medical representation learning},
  author={Ma, Chong and Jiang, Hanqi and Chen, Wenting and Li, Yiwei and Wu, Zihao and Yu, Xiaowei and Liu, Zhengliang and Guo, Lei and Zhu, Dajiang and Zhang, Tuo and others},
  journal={Advances in Neural Information Processing Systems},
  volume={37},
  pages={6126--6153},
  year={2024}
}


\end{document}